\documentclass[conference]{IEEEtran}
\IEEEoverridecommandlockouts 
\usepackage{blkarray}                                      
\usepackage{algpseudocode}                                 
\usepackage{algorithm}
\usepackage{graphicx}                                      
\usepackage{amsmath}
\usepackage{amssymb}
\usepackage{amsfonts}
\usepackage{amsthm}
\usepackage[mathcal]{eucal}
\usepackage{mathrsfs}
\usepackage{booktabs}
\usepackage{enumerate}
\usepackage{multirow}
\usepackage[subrefformat=parens,farskip=0pt,justification=centering]{subfig}
\usepackage{color}
\usepackage{cite}                                          
\usepackage{comment}                                       
\usepackage{soul}                                          
\soulregister\cite7
\soulregister\ref7
\soulregister\pageref7
\usepackage{etoolbox}                                      
\usepackage{url}
\usepackage{nth}                                           
\usepackage{bm}                                            
\usepackage{courier}
\usepackage{balance}
\usepackage{threeparttable}
\usepackage{xcolor,colortbl}
\usepackage{footnote}

\usepackage{verbatim}
\usepackage[bookmarks=false]{hyperref}
\hypersetup{
    colorlinks = true,
    citecolor  = blue,
    linkcolor  = blue,
    urlcolor   = blue,
}
\usepackage{tikz}
\usetikzlibrary{patterns,snakes}
\usetikzlibrary{positioning,calc,fit,decorations.pathmorphing,shapes.geometric, shapes.gates.logic.US, calc}
\usetikzlibrary{arrows,arrows.meta,decorations.markings,shapes,shapes.arrows}
\usetikzlibrary{decorations,decorations.pathreplacing}
\usetikzlibrary{backgrounds}
\usepackage{filecontents}                                  
\usepackage{pgfplots}
\usepackage{pgfplotstable}
\usepackage{scalefnt}
\pgfplotsset{compat=newest}
\usepackage{caption}
\usepackage{cleveref}
\Crefformat{figure}{Fig.~#2#1#3}                           
\Crefname{subfigure}{Fig.}{Figs.}
\Crefname{figure}{Fig.}{Figs.}
\Crefformat{table}{TABLE~#2#1#3}                           
\usepackage[figuresright]{rotating}

\definecolor{CUHKorange}{RGB}{244,106,18} 
\definecolor{CUHKblue}{RGB}{0,111,190}    
\definecolor{CUHKgreen}{RGB}{0,127,128}   
\definecolor{CUHKred}{RGB}{228,46,36}     
\definecolor{CUHKyellow}{RGB}{198,148,34} 
\definecolor{CUHKdark}{RGB}{114,44,114}   
\definecolor{CUHKmiddle}{RGB}{144,44,144} 
\definecolor{CUHKlight}{RGB}{167,44,167} 
\definecolor{CUHKpurple}{RGB}{117,15,109}
\definecolor{CUHKgold}{RGB}{221,163,0}
\definecolor{CUHKribbon}{RGB}{244,223,176}
\definecolor{CUHKblack}{RGB}{34,24,21}

\renewcommand{\bf}[1]{\textbf{#1}}
\renewcommand{\tt}[1]{\texttt{#1}}

\DeclareMathOperator*{\argmax}{argmax}

\usepackage{tcolorbox}
\tcbuselibrary{skins,breakable}
    {\endtcolorbox}

\usepackage[top=0.76in,bottom=0.81in,left=0.60in,right=0.60in]{geometry}
\crefname{mytheorem}{Theorem}{Theorems}
\crefname{mylemma}{Lemma}{Lemmas}
\crefname{myclaim}{Claim}{Claims}
\crefname{myproperty}{Property}{Properties}
\crefname{mycorollary}{Corollary}{Corollaries}

\algrenewcommand\textproc{\texttt}

\makeatletter
\let\OldStatex\Statex
\renewcommand{\Statex}[1][3]{%
  \setlength\@tempdima{\algorithmicindent}%
  \OldStatex\hskip\dimexpr#1\@tempdima\relax
}
\makeatother

\RequirePackage[normalem]{ulem} 
\RequirePackage{color}\definecolor{RED}{rgb}{1,0,0}\definecolor{BLUE}{rgb}{0,0,1} 

\usepackage{times}
\usepackage{algpseudocodex}
\usepackage{upgreek}
\usepackage[export]{adjustbox}

\graphicspath{{./figs/}}

\newcommand{\fname}{\texttt{LLANA} }

\newcommand{\orange}[1]{\textcolor{CUHKorange}{#1}}
\newcommand{\green}[1]{\textcolor{CUHKgreen}{#1}}

\newcommand{\blue}[1]{\textcolor{CUHKblue}{#1}}

\newcommand{\yellow}[1]{\textcolor{CUHKyellow}{#1}}

\algrenewcommand\algorithmicrequire{\textbf{Input:}}
\algrenewcommand\algorithmicensure{\textbf{Output:}}

\begin{document}
\date{}

\title{
  LLM-Enhanced Bayesian Optimization for Efficient Analog Layout Constraint Generation
}

\author{
    Guojin Chen$^{1,2}$, \quad
    Keren Zhu$^1$, \quad
    Seunggeun Kim$^2$, \quad
    Hanqing Zhu$^2$, \quad
    Yao Lai$^3$, \quad
    Bei Yu$^1$, \quad
    David Z. Pan$^2$\\
    $^1$Chinese University of Hong Kong \quad
    $^2$University of Texas at Austin \quad
    $^3$The University of Hong Kong
}



\maketitle

\begin{abstract}
Analog layout synthesis faces significant challenges due to its dependence on manual processes, considerable time requirements, and performance instability. Current Bayesian Optimization (BO)-based techniques for analog layout synthesis, despite their potential for automation, suffer from slow convergence and extensive data needs, limiting their practical application. This paper presents the \fname framework, a novel approach that leverages Large Language Models (LLMs) to enhance BO by exploiting the few-shot learning abilities of LLMs for more efficient generation of analog design-dependent parameter constraints. Experimental results demonstrate that \fname not only achieves performance comparable to state-of-the-art (SOTA) BO methods but also enables a more effective exploration of the analog circuit design space, thanks to LLM's superior contextual understanding and learning efficiency. 
The code is available at \url{https://github.com/dekura/LLANA}.
\end{abstract}
\section{Introduction}
The increasing demand for advanced analog and mixed-signal (AMS) integrated circuits (ICs) in sectors such as automotive and the Internet of Things (IoT) necessitates faster design processes and quicker time-to-market. However, the creation of analog layouts remains a predominantly manual, time-consuming, and error-prone task. Engineers rely on established practices and insights from experienced designers, incurring substantial costs and prolonging the design cycle. The growing complexities of layout-dependent effects in newer technology nodes further complicate the accurate prediction and assessment of post-layout performance. Although attempts have been made to automate the analog layout process, their integration into real-world design practices remains limited.

Existing tools for analog layout synthesis often come with significant design complexities or fail to ensure desired post-layout outcomes. Optimization-driven tools~\cite{align2019kunal} require circuit designers to manually input specific layout constraints, which are then followed during component placement and routing (P\&R). Methods relying on heuristic constraints face challenges in real-world applications due to their design-specific nature, lacking adaptability and universality across different projects. Performance-driven approaches~\cite{ou-DAC-NLPAnalogPlacement-2015} attempt to account for layout effects by deriving equations, either analytically or through sensitivity analyses. However, device miniaturization makes analytical estimations of post-layout effects increasingly inaccurate. Accurately predicting the impact of design and layout-dependent phenomena, such as mismatches and parasitics, requires extensive empirical simulations due to their increased complexity in scaled-down technologies.

Constraint generation plays a crucial role in automatic analog synthesizers~\cite{align2019kunal, Chenhao-2021CICC-Magical}, tasked with extracting physical constraints to inform and optimize the subsequent placement and routing flow. These constraints aim to minimize discrepancies between pre-layout and post-layout simulations, considering factors such as device matching, electrical current paths, and thermal management~\cite{analog-ASPDAC2016-Lin}. While recent advancements~\cite{analog_ASPDAC2022-Zhu} have focused on identifying and enforcing matching constraints, such as symmetry, an exploration into design-dependent P\&R hyperparameters, such as net weighting, remains absent.

Recent progress in analog circuit dimensioning has significantly improved the effectiveness of leveraging simulations for performance enhancement~\cite{Chenhao-2021CICC-Magical,LiuMingjie-DAC-BOMagical,ZhukerenICCAD2020,perf3dgnn-dac2024-xu,iccad22-hanqing}. These simulation-driven methods approach analog dimensioning as an optimization problem that does not reveal its internal workings, relying on circuit simulators to evaluate performance metrics. This approach requires minimal preliminary knowledge about circuit configurations, contrasting with model-dependent techniques that necessitate a deep understanding of complex design and performance interactions.
The integration of Gaussian Process (GP)-based Bayesian Optimization (BO) for automatic transistor sizing combines the strengths of both paradigms, achieving significant reductions in simulation requirements while maintaining adaptability across various circuit designs. This advancement highlights the importance of incorporating iterative simulations into the design feedback loop, inspiring the development of an integrated, feedback-oriented analog layout synthesizer.

\begin{figure}[t]
  \centering
  \includegraphics[width=0.99\linewidth]{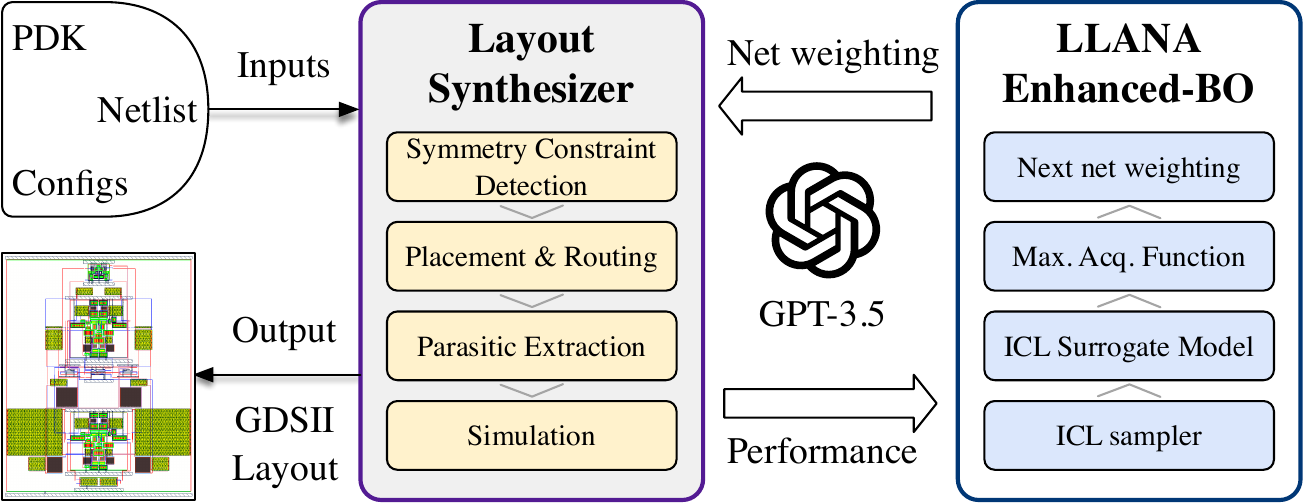}
  \caption{Overview of the \fname framework.}
  \label{fig:flowchart}
\end{figure}

Despite advances in BO-based automated analog synthesis~\cite{LiuMingjie-DAC-BOMagical}, BO faces challenges in high-dimensional tasks due to its computational demands and the complexity of probabilistic model evaluation. Its effectiveness is further limited in environments with numerous local optima, owing to its dependence on prior knowledge and the need to balance exploration and exploitation. The core of these challenges lies in accurately learning objective functions and generating solutions with minimal data, a scenario often associated with the \textit{few-shot} setting. Large Language Models (LLMs), with their extensive training on internet data, excel in \textit{few-shot} learning, demonstrating remarkable abilities in prediction, generation~\cite{brown2020language}, and contextual understanding~\cite{wei2022emergent}. This success is partially attributed to their utilization of encoded priors.

As depicted in \Cref{fig:flowchart}, this paper investigates the feasibility of leveraging LLMs to improve model-based BO for analog design-dependent parameter generation and fine-tuning, extending LLM applications beyond traditional natural language tasks. We propose a novel approach that utilizes natural language representations for BO components, aiming to harness the unique advantages of LLMs. The central inquiry of our research is whether LLMs' inherent knowledge and few-shot learning capabilities can enhance crucial aspects of BO, particularly in generating analog layout constraints, and assess the efficiency of an LLM-augmented BO pipeline functioning seamlessly from start to finish.
Our main contributions are as follows:
\begin{enumerate}
  \item We formulated the analog layout design-dependent parameter space and provided a benchmark for BO based on Gaussian processes.
  \item For the first time, we applied LLM-enhanced BO to analog design-dependent constraint generation, exploring the capabilities of LLM-based design space exploration.
  \item All the algorithms, experiments and benchmarks are open-sourced at \url{https://github.com/dekura/LLANA}.
\end{enumerate}

\section{Preliminaries and Background}
\subsection{Analog placement}
Analog circuit layout synthesis involves placement and routing for both building and macro blocks. Placement optimizes component locations within a predefined bounding box to minimize wire lengths, while routing finalizes the layout using established component positions, pin locations, and connection data. Symmetry constraints are respected to minimize mismatches. The placement and routing engine is design-independent, ensuring efficiency and coherence. The objective function $f$ for analytical global placement is defined as~\cite{magical29xu}:
\begin{equation}
f = \sum_i \alpha_i \cdot f_{WL_i} + \beta \cdot f_{AREA} + f_{other},
\label{eq:ana_obj}
\end{equation}
where $f_{WL_i}$ is the wirelength of net $i$, $f_{AREA}$ is the total layout area, $\alpha_i$ and $\beta$ are the net weightings and area minimization factor, and $f_{other}$ represents other necessary objectives.

\subsection{Analog design-dependent parameter constraint:}
\label{subsec:design_params}
Analog design-dependent parameter constraints, along with symmetric constraints, define the layout and efficiency of integrated circuits. Net weightings ($\alpha_i$ in \cref{eq:ana_obj}) direct global placement strategies, affecting the floorplan and performance. Higher weights reduce parasitics by aligning connected components more closely. Wire widths and routing sequences manage the trade-off between parasitic resistance and capacitance, determining layout complexity. Hyperparameters, such as the area minimization factor ($\beta$ in \cref{eq:ana_obj}) and net spacing, impact placement and routing optimization, reducing coupling interference. Design-dependent parameters and hyperparameters are crucial for optimizing analog circuit performance and layout. This paper focuses on the impact of \textbf{net weightings} $x_i$ on layout performance.
\section{Algorithm and Framework}

\subsection{BO-based methods as baselines}

Prior works of analog constraint generation have primarily focused on multi-objective Bayesian optimization(MOBO) and using Gaussian processes(GP) as surrogate model~\cite{LiuMingjie-DAC-BOMagical}, formulated as:
\begin{equation}
  \min (f_1(x), f_2(x), \ldots, f_m(x)),
  \label{eq:mobo_obj}
\end{equation}
where $f_i(x)$ represents the post layout performance metric obtained through simulation after automatic layout generation using Magical~\cite{magical29xu},
and $x \in \mathbb{R}^d$ are the optimal design specific layout parameters introduced in \Cref{subsec:design_params} that minimize the performance metrics.

\begin{algorithm}[h]
  \caption{MOBO~\cite{LiuMingjie-DAC-BOMagical}}
  \footnotesize
    \begin{algorithmic}[1]
        \Require Sampled data points $x^t$, $\{f_i(x^t)\}_{i = 1}^m$
        \Ensure Next net weighting $x_{t+1}$
        \State Initialize Pareto set $P$ to $\emptyset$;
        \Function{MOBO}{$x^t$, $\{f_i(x^t)\}_{i = 1}^m$}
            \If{$t < N_{\text{random}}$}
                \State Random sample $x_{t+1}$ from the design space;
            \Else
                \State Update $P$ with $x^t$ and reference point $r$;
                \State Update GP models with $f_i(x)$;
                \State Update GP model with $x^t$, $\{f_i(x^t)\}_{i = 1}^m$;
                \State Optimize acquisition function to obtain $x_{t+1}$;
            \EndIf
            \State \Return $x_{t+1}$;
        \EndFunction
    \end{algorithmic}
  \label{alg:mobo}
\end{algorithm}

As presented in \Cref{alg:mobo}, MOBO iteratively samples data points $x^t$ and their corresponding performance metrics $\{f_i(x^t)\}_{i = 1}^m$ to update the GP models and optimize the acquisition function to obtain the next net weighting $x_{t+1}$.

\subsection{Large Language Models (LLMs)-enhanced BO}
BO's effectiveness depends on the quality of \textbf{surrogate models} and \textbf{sampling strategies}, challenged by limited data, sensitivity to inaccuracies, and the difficulty of incorporating \textbf{prior knowledge} into new tasks. Recently, there has been a growing trend of utilizing AI and LLMs to address challenges within the EDA domain~\cite{lai2024analogcoder, lai2024scalable, ren2022machine,pujar23automated,chateda23he,Zhao_2024_CVPR, zhao2022end, ICCAD20_DAMO,blocklove2023chip,ICCAD22_LayoutTransformer,ICCAD22_AdaOPC,thakur2023verigen,DAC23_DiffPattern,liu2023rtlcoder,ICCAD21_DevelSet,ispd24he,DAC23_Nitho,liu2023chipnemo,ICCAD23:AlphaSyn,liu2023verilogeval,TCAD_DiffSMO,lu2023rtllm,DAC24_BiSMO,chen2024dawn,L2OILT_TCAD,circuitops,ISPD21_Pointcloud,delorenzo2024make,chipformer-icml23-lai,NEURIPS2022_maskplace}. \cite{xie2021explanation} explained LLM in-context learning (ICL) as performing implicit Bayesian inference~\cite{brown2020language}. LLMs can enhance BO by leveraging: (1) prior knowledge through ICL for tapping into pre-trained insights~\cite{mirchandani2023large}, (2) the ability to generalize from limited examples, aiding in efficient exploration~\cite{liu2024large}, and (3) processing contextual information to enrich optimization tasks and search strategies~\cite{yang2023large}.

\subsection{LLM-enhanced BO Framework}
\textbf{LLM-enhanced Initial Design}: As illustrated in \Cref{fig:initial_design_prompt}, by leveraging LLM's `\textit{zero-shot}' capabilities, we can generate better initial designs through prompt engineering.
\begin{figure}[h]
  \centering
\begin{tcolorbox}[colback=gray!8, colframe=gray!70, width=1.\linewidth, size=small]
  \footnotesize
Assist me with automated machine learning using \{\orange{model}\}. Explore these hyperparameters: \{\orange{configurations, type, ranges}\}. Suggest \{\green{number of recommendations}\} diverse, effective configs for BO hyperparameter tuning, without ``None''. Respond with an un-enumerated list of dictionaries, each describing a recommended config.
\end{tcolorbox}
\caption{Prompt for initial design generation}
  \label{fig:initial_design_prompt}
\end{figure}

\textbf{LLM-enhanced surrogate modeling}: BO constructs a surrogate model $p(f|x)$ using $m$ observed input-output pairs $\mathcal{X}_m:=\{(x_i, f_i)\}_{i=1}^m$. Common models include the GP~\cite{LiuMingjie-DAC-BOMagical} and random forests (SMAC)~\cite{lindauer2022smac3}. The \fname framework serializes the optimization trajectory into natural text, e.g., for an RF model, `[max\_depth is 15, min\_samples\_split is 0.5, ..., performance is 0.9]'. These text representations, denoted as $\mathcal{D}_m$, along with the problem description and queried few-shot examples $x_k^{m1}$, are input to the LLM, denoted as $\mathcal{D}_m^{m1}$. The LLM outputs a predicted score and associated probability: $(\hat{f}_k, p(\hat{f}_k)) =$ \fname$(x_k^{m1}, \mathcal{D}_m^{m1})$. A shuffling mechanism randomly permutes the few-shot examples within $\mathcal{D}_m^{m1}$ to enhance robustness. The in-context learning prompts for the LLM-enhanced \textit{surrogate model} are shown in \Cref{fig:disc_prompt}.

\begin{figure}[H]
    \centering
\begin{tcolorbox}[colback=gray!8, colframe=gray!70, width=.99\linewidth, size=small]
  \footnotesize
  The following are examples of the performance of a \{\orange{model}\} measured in \{\orange{metric}\} and the corresponding model hyperparameter configurations. The model is evaluated on a tabular \{\orange{task}\} containing \{\yellow{number of classes}\}. The tabular dataset contains \{\yellow{number of samples}\} samples and \{\yellow{number of features}\} features (\{\yellow{number of categorical features}\} categorical, \{\yellow{number of continuous features}\} numerical).\\
  Your response should only contain the predicted accuracy in the format \#\# performance\#\#.\\
  Hyperparameter configuration: \{\blue{C1}\}.
  Performance: \{\blue{P1}\}. \\
  ...\\
  Hyperparameter configuration: \{\blue{Cn}\}.
  Performance: \{\blue{Pn}\}. \\
  Hyperparameter configuration: \{\blue{configuration to predict}\}.
  Performance:\{\}.
\end{tcolorbox}
\caption{Prompt for \textbf{discriminative surrogate model}. }
\label{fig:disc_prompt}
\end{figure}

\textbf{Acquisition strategies}:
Sampling candidate points is crucial in BO, as high-potential points can accelerate convergence to the optimum. \fname introduces a novel mechanism to conditionally generate candidate points based on desired objective values through ICL. The samples are generated from high-potential regions by conditioning on a desired objective value $x^{\prime}: \tilde{h}_m \sim p(f | x^{\prime};\mathcal{D}_m)$, leveraging the few-shot generation capabilities of LLMs.
The desired objective value is defined as $x^{\prime}\!=\!x_{\min}-\alpha\!\times\!(x_{\max}-x_{\min})$, where $x_{\max}$ and $x_{\min}$ are the worst and best observed points at related objective values, and $\alpha$ is an exploration hyperparameter. Positive $\alpha$ sets $x^{\prime}$ to improve over $x_{\min}$, while negative $\alpha$ results in a more conservative target value within the observed range. We implement $p(f | x^{\prime}; \mathcal{D}_m)$ through ICL, generating $M$ candidate points independently, i.e., $\tilde{h}_k \sim \operatorname{\fname}(x^{\prime}, \mathcal{D}_m^{m1})$, and select the point that maximizes the acquisition function as the next point to evaluate, using a sampling-based approximation to optimize the acquisition function.
\begin{figure}[h]
  \centering
\begin{tcolorbox}[colback=gray!8, colframe=gray!70, width=.99\linewidth, size=small]
  \footnotesize
  The following are examples of the performance of a \{\orange{model}\} in \{\orange{metric}\} and corresponding hyperparameter configs,
  evaluated on a tabular \{\orange{task}\} task with \{\yellow{number of classes}\} classes, \{\yellow{number of samples}\} samples, 
  \{\yellow{number of features}\} features (\{\yellow{number of categorical features}\} categorical, 
  \{\yellow{number of continuous features}\} numerical). Hyperparameter ranges: \{\orange{configuration and type}\}. \\
  Recommend a config to achieve \{\blue{target score}\}, avoiding min/max/rounded values, using highest precision.
  Respond with only the predicted config, as \#\# configuration \#\#.\\
  Performance: \{\blue{P1}\}. Hyperparameter config: \{\blue{C1}\}\\
  ...\\
  Performance: \{\blue{Pn}\}. Hyperparameter config: \{\blue{Cn}\}\\
  Performance: \{\blue{performance used to sample configuration}\}\\
  Hyperparameter config:\{\}.
\end{tcolorbox}
\caption{Prompt for \textbf{candidate sampling}.}
  \label{fig:acq_prompt}
\end{figure}

\textbf{End-to-end \fname framework}:
The end-to-end procedure iteratively performs three steps, as depicted in \Cref{alg:end-to-end-icl}.
(1) sample $M$ candidate points $\{\tilde{h}_m\}_{m=1}^M$ through ICL.
(2) evaluate $M$ points using the ICL surrogate model, i.e. $p(f \mid \tilde{h}_m)$ to obtain scores $\{a(\tilde{h}_m)\}_{m=1}^M$ according to an acquisition function.
(3) select point with the highest score to evaluate next, $h=\argmax_{\tilde{h} \in\{\tilde{h}_m\}_{m=1}^M} a(\tilde{h})$.
In \fname, we use expected improvement (EI), $a(\tilde{h}_m)=\mathbb{E}[\max (p(f \mid \tilde{h}_m)-f(h_{\text {best }}), 0)]$.
\begin{algorithm}[h]
  \caption{End-to-end \texttt{LLANA}-BO with ICL}
  \footnotesize
  \begin{algorithmic}[1]
  \Require Initial best point $h_{\text{best}}$
  \Ensure Optimal point $h^{\star}$
  \Function{\texttt{LLANA}}{$h_m$}
  \State Sample $M$ candidate points $\{\tilde{h}_m\}_{m=1}^M$ through \textit{ICL};
  \State Evaluate $p(f\mid \tilde{h}_m)$ using the \textit{ICL} surrogate model;
  \State EI score $a(\tilde{h}_m) = \mathbb{E}[\max(p(f \mid \tilde{h}_m) - f(h_{\text{best}}), 0)]$;
  \State $h=\argmax_{\tilde{h} \in\{\tilde{h}_m\}_{m=1}^M} a(\tilde{h})$;
  \If{$f(h) > f(h_{\text{best}})$}
  \State $h_{\text{best}} = h$;
  \EndIf
  \State \Return $h_{\text{best}}$ as $h$;
  \EndFunction
  \end{algorithmic}
  \label{alg:end-to-end-icl}
\end{algorithm}
\section{Experiments}
\label{sec:exp}
The \fname framework is implemented in Python, utilizing the `gpt-3.5-turbo' model from OpenAI with hyperparameters set to $\alpha = -0.1$ and $M = 20$. The designs used in the experiments are two-stage operational amplifiers from ~\cite{LiuMingjie-DAC-BOMagical}, and the performance benchmark is evaluated by Cadence Spectre after layout generation using Magical~\cite{magical29xu}. The design contains a total of 36 devices with 14 design-dependent parameters of the critical \textbf{net weighting} selected by the MOBO algorithm introduced in \Cref{alg:mobo}. A dataset of 500 design-performance pairs is prepared, with 400 pairs used for training and 100 pairs for testing. The optimization objectives chosen are the common-mode rejection ratio (CMRR) and absolute input-referred offset (Offset) voltage. The experiments include three ML models: RandomForest (RF), AdaBoost, and DecisionTree. The scoring function is the mean squared error (MSE), and the acquisition function is expected improvement (EI).

\subsection{Evaluation of the surrogate model}
The performance of the \fname framework is first evaluated against single-objective GP and SMAC~\cite{lindauer2022smac3} on both CMRR and Offset datasets. For GP and SMAC, the average result of three ML models is used. The performance of the surrogate model is assessed using prediction performance and uncertainty calibration metrics. The Normalized Root Mean Square Error (NRMSE) and the coefficient of determination ($R^2$ score) of the prediction on the test set are used as performance metrics. Calibration is evaluated using the log predictive density (LPD) and the normalized regret.
The regret metric is defined as
$\min_{h \in \mathcal{H}_t}(f(h)-f_{\text{min}}^*) / (f_{\text{max}}^*-f_{\text {min }}^* )$, where $\mathcal{H}_t$ denotes the points chosen up to trail $t$, and $f_{\text{max}}^*$,$f_{\text {min }}^*$ represent the best and worst scores~\cite{lindauer2022smac3}.

\begin{figure}[tbp!]
    \centering
    \includegraphics[width=.95\linewidth]{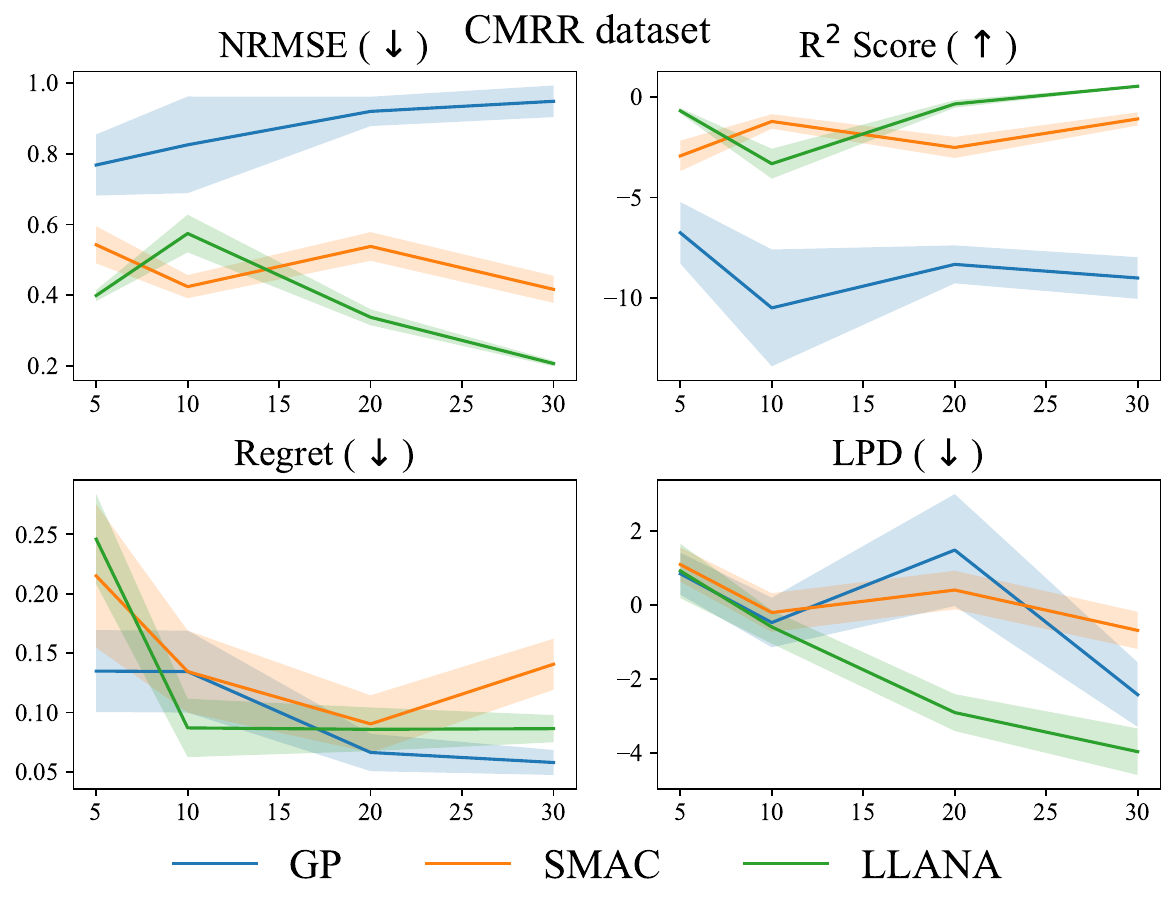}
    \caption{Comparison of \fname, GP, and SMAC~\cite{lindauer2022smac3} on CMRR dataset. The x-axis is the number of observed points.}
    \label{fig:cmrr_score}
\end{figure}

\Cref{fig:cmrr_score} plots the performance of \fname against GP and SMAC on the CMRR dataset. \fname outperforms GP and SMAC in terms of NRMSE and $R^2$ score. Moreover, this trend becomes more apparent when there are more observed points. As for normalized regret, \fname attains lower regret than GP and SMAC when $n < 15$. For uncertainty calibration, \fname achieves lower LPD than GP and SMAC, producing the best uncertainty quantification.

\begin{figure}[t]
  \centering
  \includegraphics[width=.95\linewidth]{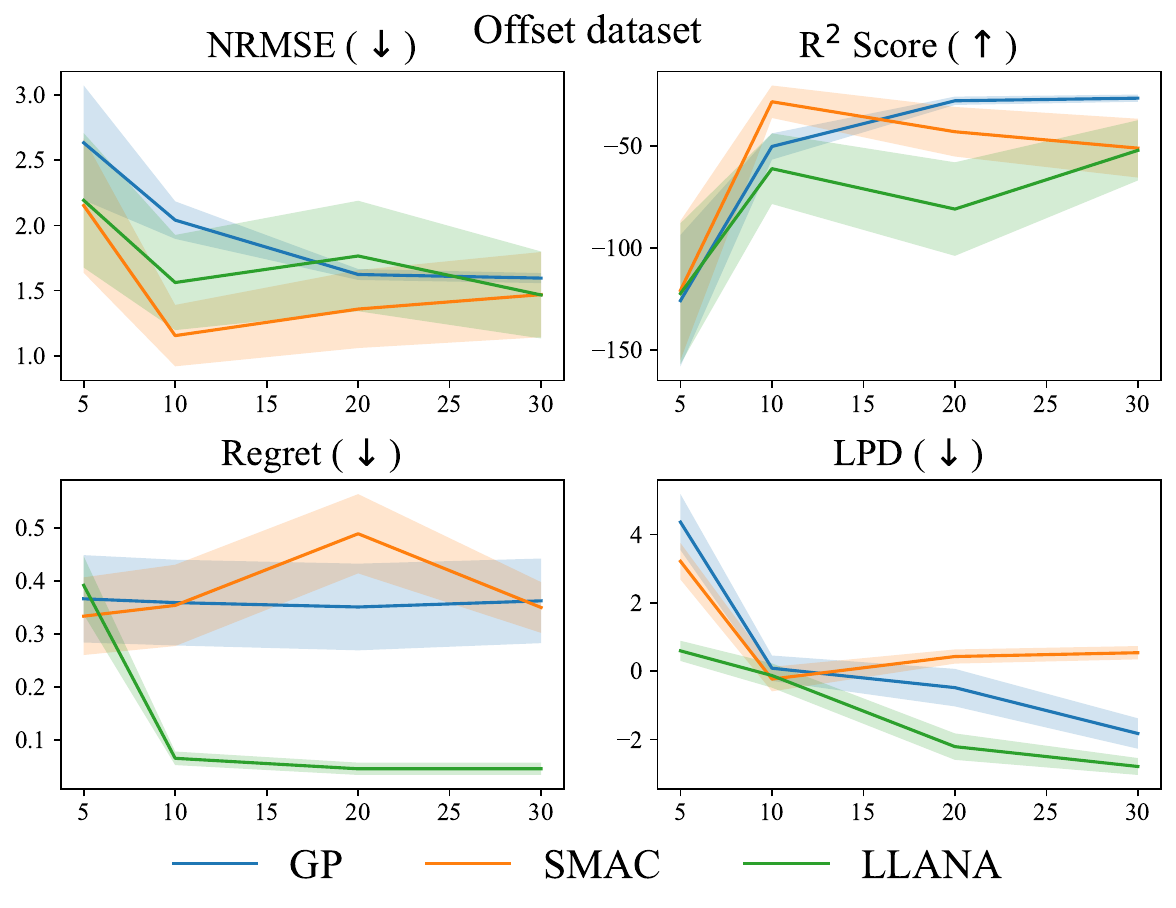}
  \caption{Comparison of \fname, GP, and SMAC~\cite{lindauer2022smac3} on Offset dataset. The x-axis is the number of observed points.}
  \label{fig:offset_score}
\end{figure}

\Cref{fig:offset_score} shows the performance of \fname against GP and SMAC on the Offset dataset. This time, as the number of observed points increases, \fname can achieve NRMSE and $R^2$ scores close to those of GP and SMAC when $n = 30$. However, the regret and LPD of \fname consistently remain far ahead of GP and SMAC. Additionally, we observe that, consistent with prior findings, \fname excels in earlier stages of the search, when fewer observations are available. As such, empirical evidence supports that permuting few-shot examples, while straightforward in implementation, improves both uncertainty quantification and prediction performance, both critical aspects of balancing exploration and exploitation.
\fname performs effectively as an end-to-end pipeline, exhibiting sample-efficient search. Its modularity further enables individual components to be integrated into existing frameworks. Surrogate models implemented through ICL can produce effective regression estimates with uncertainty, although there is a tradeoff of stronger prediction performance with worse calibration than probabilistic methods. The LLM's encoded prior is crucial to improving the efficacy of such surrogate models.

\section{Conclusion and Discussion}
This work introduces \fname, a novel framework that integrates LLMs with BO to address the challenge of generating analog net weighting constraints. The approach incorporates two key enhancements: surrogate models of the objective function through ICL and a candidate point sampler capable of conditional generation for specific target values. The study on the analog constraint problem reveals performance improvements across the three integrations, particularly when working with limited sample sizes. Moreover, \fname demonstrates potential as a stand-alone BO method, exhibiting slightly better results on CMRR and Offset benchmarks compared to existing techniques. However, further research is necessary to fully assess the extent of \fname's capabilities and its potential for generalization across a broader range of optimization tasks. Codes and experiment results are available at \url{https://github.com/dekura/LLANA}.

\textbf{Limitations and future work.} Despite \fname's higher computational cost due to LLM inference, the findings suggest that it trades off computational complexity for improved sample efficiency, which is essential in black-box optimization. This indicates the potential for combining \fname with more computationally efficient methods to achieve better solutions. Furthermore, unlike the multi-objective approach in \cite{LiuMingjie-DAC-BOMagical}, this work focuses solely on single-objective BO. A promising avenue for future research is extending \fname to handle multi-objective and higher-dimensional BO tasks with more complex search spaces, enhancing its applicability and impact in optimization.
{
\bibliographystyle{IEEEtran}
\bibliography{ref/Top,ref/bench,ref/analog,ref/llm,ref/ai}
}

\end{document}